\documentclass{article}
\usepackage{spconf,amsmath,graphicx,hyperref}
\usepackage{booktabs}
\usepackage{tipa}
\usepackage{multirow}

\title{Quantifying Speaker Embedding Phonological Rule Interactions \\ in Accented Speech Synthesis}
\name{Thanathai Lertpetchpun$^{1*}$,\, Yoonjeong Lee$^{1*}$,\, Thanapat Trachu$^2$,\, Jihwan Lee$^1$, \\ \textit{Tiantian Feng}$^1$,\, \textit{Dani Byrd}$^3$,\, \textit{Shrikanth Narayanan}$^{1,2,3}$\thanks{*Equal contribution}}
\address{$^1$Signal Analysis and Interpretation Lab, University of Southern California \\
$^2$Thomas Lord Department of Computer Science, University of Southern California \\
$^3$Department of Linguistics, University of Southern California 
}%
\usepackage{fancyhdr}
\fancypagestyle{firstpage}{
\fancyhf{}
\fancyfoot[C]{\footnotesize Copyright 2026 IEEE. Published in ICASSP 2026 - 2026 IEEE International Conference on Acoustics, Speech and Signal Processing (ICASSP), scheduled for 3-8 May 2026 in Barcelona, Spain. Personal use of this material is permitted. However, permission to reprint/republish this material for advertising or promotional purposes or for creating new collective works for resale or redistribution to servers or lists, or to reuse any copyrighted component of this work in other works, must be obtained from the IEEE. Contact: Manager, Copyrights and Permissions / IEEE Service Center / 445 Hoes Lane / P.O. Box 1331 / Piscataway, NJ 08855-1331, USA. Telephone: + Intl. 908-562-3966.}

}
\begin{document}
\ninept
\maketitle

\begin{abstract}


Many spoken languages, including English, exhibit wide variation in dialects and accents, making accent control an important capability for flexible text-to-speech (TTS) models. Current TTS systems typically generate accented speech by conditioning on speaker embeddings associated with specific accents. While effective, this approach offers limited interpretability and controllability, as embeddings also encode traits such as timbre and emotion. In this study, we analyze the interaction between speaker embeddings and linguistically motivated phonological rules in accented speech synthesis. Using American and British English as a case study, we implement rules for flapping, rhoticity, and vowel correspondences. We propose the phoneme shift rate (PSR), a novel metric quantifying how strongly embeddings preserve or override rule-based transformations. Experiments show that combining rules with embeddings yields more authentic accents, while embeddings can attenuate or overwrite rules, revealing entanglement between accent and speaker identity. Our findings highlight rules as a lever for accent control and a framework for evaluating disentanglement in speech generation.

\end{abstract}
\section{Introduction}

Accents are a defining source of variation in spoken language, and English provides a striking case. Spoken by approximately one-fifth of the world's population, yet with only one-fourth native speakers~\cite{crystal2003english,ethno2024english}, the English language has developed a broad range of accent varieties from both native and non-native speaker backgrounds~\cite{wells1982accents}. These include regional varieties such as American, British, and Australian English, as well as global variants such as Indian English. ~\cite{kachru1990world,schneider2007postcolonial}. Capturing such variation has become an important goal for speech generation technologies. 

Accent in TTS is typically controlled by conditioning on speaker embeddings \cite{zhou2024multi, zhou2024accented}. However, embeddings also encode multiple traits unrelated to the accent, such as voice timbre \cite{du2023speaker, lee23f_interspeech}, speaker emotion \cite{cho2025diemo, lertpetchpun2023instance, melstts}, and background noise \cite{fujita2024noise}, making accent representation opaque and difficult to control. To address this, we use linguistically motivated phonological rules as targeted probes of accent control. Our goal is not to replace data-driven models with synthesis-by-rule but to test how explicit phoneme-level transformations interact with speaker embeddings in modern TTS systems.

We focus on three well-documented phonological processes that differentiate American and British English varieties: flapping, rhoticity, and vowel correspondences. In American English, intervocalic \textipa{/t/} is often realized as a flap \textipa{[R]} in unstressed contexts (e.g., \textit{water} $\rightarrow$ \textipa{[waR\textrhookschwa]}), while British English typically retains \textipa{[t]}. Rhoticity also diverges: American English is rhotic, preserving post-vocalic \textipa{\*r} (e.g., \textit{car} $\rightarrow$ \textipa{[ka\*r]}), whereas most British varieties are non-rhotic, deleting or vocalizing \textipa{\*r} in coda position (e.g., \textit{car} $\rightarrow$ \textipa{[kA:]}). Finally, systematic vowel correspondences arise across lexical sets. For example, \textit{bath} is pronounced as \textipa{/b\ae T/} in American English but \textipa{/bAT/} in British English, and \textit{goat} as \textipa{[goUt]} versus \textipa{[g@Ut]}.

These selected contrasts are not modeled to capture every nuance of accent variation. Instead, we deliberately operationalize them as big-stroke transformations—salient enough to shift perceived accent without overfitting dialectal micro-variation. This makes them a useful testbed for probing the balance between embedding-driven and linguistic control, focusing less on mimicking full dialects and more on revealing where embeddings and linguistic structure interact. Our substitutions target robust cross-accent pronunciation patterns that have been consistently documented in linguistic descriptions~\cite{wells1982accents,trudgill1999dialects} and behavioral studies of accent perception~\cite{flege1995sll,clopper2004homebodies}, making them ideal probes for accent strength and a principled lens on disentanglement in speech generation.

To quantify accent strength we use an accent classification model, where a higher predicted probability for a given accent indicate stronger accentedness. A caveat in this approach, however, is that predicted probabilities are inherently tied to the model's training task and the particular accent contrasts it encodes, making them task-specific rather than general measures. To complement this, we also compute accent embedding similarity. If two utterances yield embeddings that lie close together in the accent space, they are assumed to share similar accent characteristics. 

While useful, neither of these measures directly captures the impact of phonological rules. Our rules are deliberately coarse, targeting the most salient cross-accent shifts, but the actual realization may be moderated by the speaker embedding. To evaluate this interaction, we introduce the phoneme shift rate (PSR), a novel metric that quantifies how much speaker embeddings preserve or overwrite the rule-driven phoneme mappings. For example, when we convert \textipa{[l\ae R@\*r]} (\textit{latter}) into a more British-like target \textipa{[lAt@@]}, the synthesized output may still realize \textipa{[l\ae t\textrhookschwa]}, reflecting a partial pull back toward the American form. 
PSR measures such cases of gradient reinforcement or attenuation, offering a direct probe of where rule-based transformations hold versus where embeddings dominate. We view PSR as a first step toward systematic evaluation of disentanglement in TTS, offering an interpretable way to steer accent strength that is linguistically grounded yet flexible enough for modern TTS.

The contributions of this study are: 
1) We introduce a small set of linguistically guided phonological rules for highly salient dialect differences that transform accents between American and British English, offering a controlled and interpretable probe for accent in TTS.
2) We introduce the phoneme shift rate (PSR), a novel metric to evaluate how rule-based transformations are preserved or attenuated by speaker embeddings. 
3) We present a systematic analysis of the interplay between phonological rules and speaker embeddings, showing how coarse, knowledge-driven constraints can reveal the degree of disentanglement in speech generation. 
Together, these results demonstrate that knowledge-driven rules, even at a coarse granularity, can be effective levers for accent control, providing interpretable guidance within modern data-driven TTS systems. Speech samples are available\footnote{\url{https://sav-eng.github.io/icassp_samples.html}}.
Code implementing the phonological rules is in https://github.com/linguistylee/KAtDial.

\vspace{-1mm}
\section{Phonological Rules}

We define three sets of substitution rules to map American to British English phoneme sequences.
%

\vspace{1.5mm}
\noindent \textbf{Flapping:}  Intervocalic \textipa{/t/} realized as \textipa{[R]} in American English is mapped to [t] in British English (e.g., \textit{water}: \textipa{[waR\textrhookschwa]} $\rightarrow$ \textipa{[wAt@]}).

\vspace{1.5mm}
\noindent \textbf{Rhoticity:} 
Post-vocalic \textipa/{\*r}/ is retained in American English but deleted or vocalized in British English (e.g., \textit{car}: \textipa{[\textipa{ka\*r}]} $\rightarrow$ \textipa{[kA:]})

\vspace{1.5mm}
\noindent \textbf{Vowel correspondences:} 
Systematic mappings are applied across lexical sets, such as TRAP, BATH, and GOAT (e.g., bath: \textipa{/b\ae T/} $\rightarrow$ \textipa{/bAT/}; goat: \textipa{[goUt]} $\rightarrow$ \textipa{[g@Ut]}).

The rules are applied as one-to-one substitutions, with the phoneme character count strictly preserved across the American and British IPA sequences, ensuring that differences in synthesized accent strength arise only from segmental mappings and speaker embeddings. Although this work focuses only on the mapping between American to British, additional accents can be supported by defining corresponding rule sets and applying the same pipeline.

\vspace{-1mm}
\section{Application to Speech Generation Tasks}

\begin{figure}[h!]
    \centering
    \vspace{-5mm}
    \includegraphics[width=0.40\textwidth]{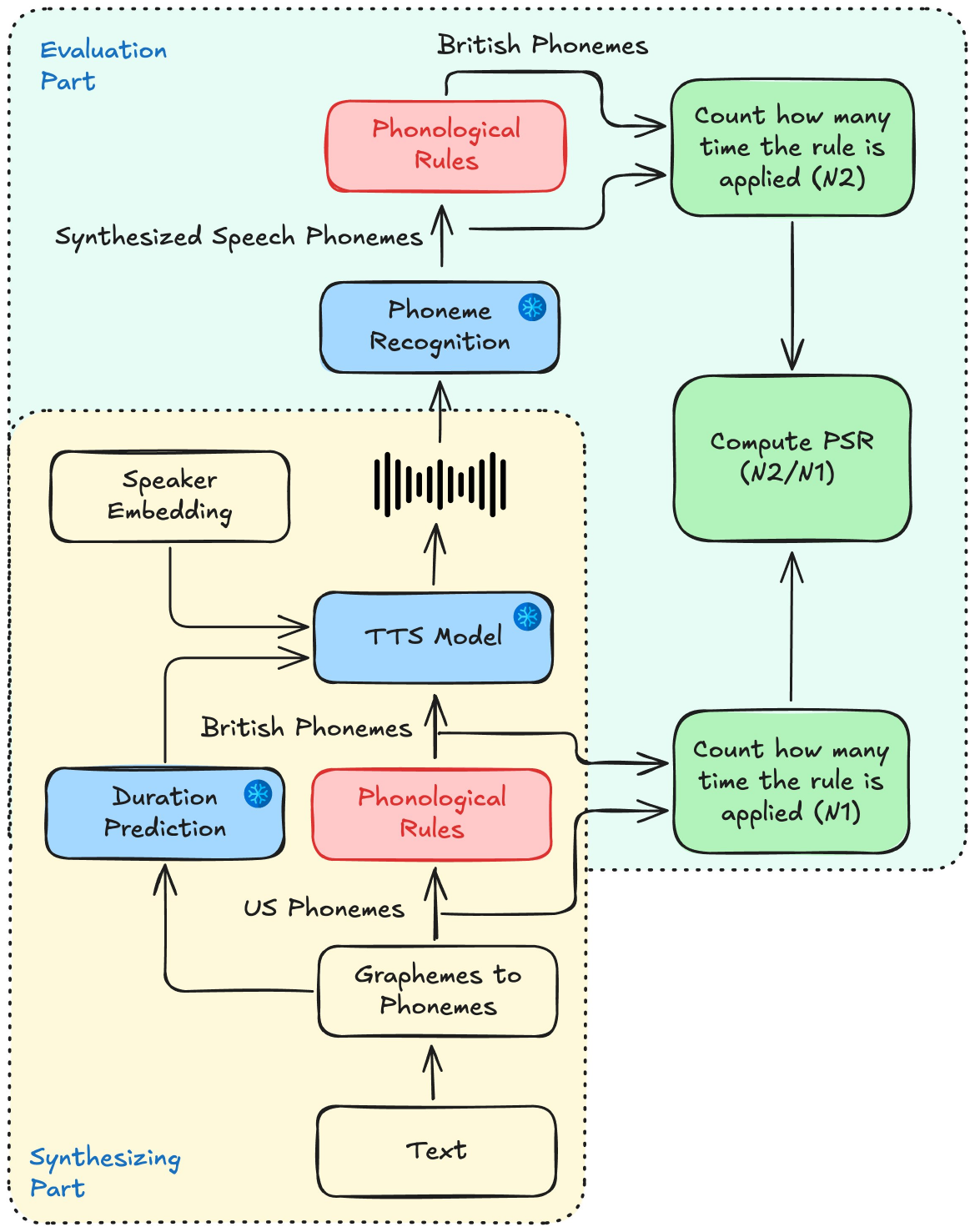}
    \vspace{-10mm}
    \caption{Synthesis and evaluation pipeline. Inputs to Kokoro TTS are a phoneme sequence (American or British), a speaker embedding, and fixed phoneme durations. The yellow path shows rule-based phoneme substitutions prior to synthesis, and the green path shows evaluation of their effect on the synthesized accented speech.}
    \vspace{-5mm}
    \label{fig:diagram}
\end{figure}

To implement our phonological rules, we follow the pipeline illustrated in Fig~\ref{fig:diagram}: 1) obtain an American English phoneme sequence from normalized text using Misaki G2P~\footnote{\url{https://github.com/hexgrad/misaki}}; 2) apply the proposed phonological rules to derive the corresponding British phoneme sequence; 3) synthesize speech outputs from both inputs.

Speech is generated using Kokoro TTS~\footnote{\url{https://github.com/hexgrad/kokoro}}. Inputs to the model include the phoneme sequence (American or British), a speaker embedding, and fixed phoneme durations. The phoneme character count and duration are strictly preserved across all conditions, ensuring that observed accent differences reflect the interaction of phonological substitution rules with speaker embeddings rather than confounds from timing or text normalization. 

\vspace{-1mm}
\section{Metrics}
We evaluate our approach along two dimensions: accent strength and naturalness. For indexing accent strength, we use two models: Vox-Profile \cite{feng2025vox} and Phoneme Recognition \cite{xu2021simple} (Wav2Vec2Phoneme). Vox-Profile  evaluates the synthesized speech with regard to the accent classifier output; while Phoneme Recognition measures how phoneme realizations shift under the influence of speaker embedding. The evaluation metrics are explicated further below.

\vspace{-2mm}
\subsection{Vox-Profile Accent Classifier}
Vox-Profile~\cite{feng2025vox} is a benchmark tool that predicts multiple speaker traits, including speaker accent. In this work, we use its Whisper-based broad accent classifier, trained on more than 100k unique utterances from 11 diverse data sources.
Specifically, the classifier assigns each input utterance to one of three accent categories: North American English, British Isles English, and English spoken with other language backgrounds. To measure the accentedness in synthesized speech, we adopt two complementary measures, following prior work~\cite{wang2025maskgct, zhong2025pairwise, zhong2025accentbox, inoue2025macst}. The first is the \textbf{accent probabilities}: logits from the classifier's final layer are converted to class probabilities with the softmax function, and the average probability assigned to the target accent (North American or British Isles) is used as the index of accent strength. The second is \textbf{accent similarities:} cosine similarity is computed between the accent embedding of each synthesized speech and a group-level reference accent embedding. The reference group-level embedding is derived by averaging accent embeddings from real speech of each accent in the Vox-Profile test set.

\subsection{Phoneme Shift Rate (PSR)}
\label{sec:PSR}
Speaker embeddings encode many dimensions of speaker traits, and therefore condition speech generation on them. They can influence not only timbre but also phoneme realization. To quantify this interaction with our phonological rules, we introduce a novel metric called phoneme shift rate (PSR). 

Without loss of generosity, we use the example in generating the British target speech described earlier. In this case, PSR is defined as the ratio $\frac{N2}{N1}$, where $N1$ is the number of phoneme substitutions specified by our rules to convert a US phoneme sequence into a British target, and $N2$ is the number of substitutions that must still be applied when re-applying the same rules to the phoneme transcript of the synthesized British output. If the output perfectly respects the rule-based transformations, $N2 = 0$ and $PSR = 0$. If the speaker embedding completely overrides the rule inputs, $N2=N1$ and $PSR = 1$, under the assumption that TTS and phoneme recognition have perfect accuracy.

Crucially, while our rules are modeled as categorical substitutions (e.g., mapping \textipa{/t/} to \textipa{[R]}, deleting post-vocalic \textipa{/\*r/}, or shifting \textipa{/ae/} to \textipa{A}), phonological categories are realized gradiently in natural speech. The same holds in synthesis: a rule may surface fully, partially, or not at all, depending on how strongly the speaker embedding shapes the acoustic outcome. For example, a British target like \textipa{[lAt@:]} (\textit{latter}) may surface as \textipa{[l\ae t\textrhookschwa]}, reflecting a partial drift back toward the American form.

PSR captures this continuum, quantifying how rule-based transformations interact with the gradient pressures of speaker embeddings. It thus goes beyond ``rule errors'' providing a diagnostic for accent control and a novel framework for analyzing how linguistic rules and embeddings compete or reinforce one another in synthesis.

\subsection{Naturalness Evaluation with UTMOS}
We also evaluate the naturalness of synthesized speech to ensure that applying phonological rules shifts accent strength without degrading the naturalness of the output speech. For this, we use UTMOS~\cite{saeki2022utmos}, a non-intrusive model trained to estimate human naturalness ratings (MOS) of speech without requiring subjective listening tests. UTMOS outputs a score ranging from 1 (least natural) to 5 (human-level naturalness). We compute UTMOS for each utterance and average across the set to obtain an overall naturalness measure.

\section{Datasets and Experimental Setups}
We use a pretrained TTS model, Kokoro-82M v0.19\footnote{The model is available at \url{https://huggingface.co/hexgrad/Kokoro-82M}.}, a multilingual TTS model supporting eight languages including English, Japanese, and Mandarin. The model takes a speaker embedding and a phoneme sequence as input. For English, 28 preset speaker embeddings are available: 20 are derived from American English speakers, and 8 are from British speakers. Our experiments manipulate both the speaker embeddings and phonological rules to control accent strength. Synthesized utterances are based on transcripts from LibriTTS-R \cite{koizumi2023libritts} on the train-clean-100 subset. For consistency, we use \textit{af\_heart} as the primary American speaker, \textit{bm\_fable} as the primary British speaker and assign the duration of each phoneme from \textit{af\_heart }uniformly to British speakers. In total, the synthesized speech spans 33k utterances and 55.4 hours of speech.

\subsection{Rule Application Configurations}
To assess the contribution of individual phonological rules, we conduct analyses under three configurations: (1) speaker embeddings alone, (2) embeddings plus a single transformation rule, and (3) embeddings plus the full set of rules. For the full set, we also perform ablation experiments in which one rule is removed at a time. This design allows us to isolate the effect of each rule on accent strength while also evaluating the combined benefit of data-driven and rule-driven approaches.

\section{Results and Discussion}
We present results from three sets of experiments. First, we fix the speaker embedding input to the TTS and vary the number of applied rules to assess how each rule contributes to accented speech. Second, we analyze synthesized phoneme sequences to see how speaker embeddings interact with or override rule-based transformations. Finally, we examine the role of individual speaker embeddings in shaping accented output.

\begin{figure*}[h!]
    \centering
    \includegraphics[width=\linewidth]{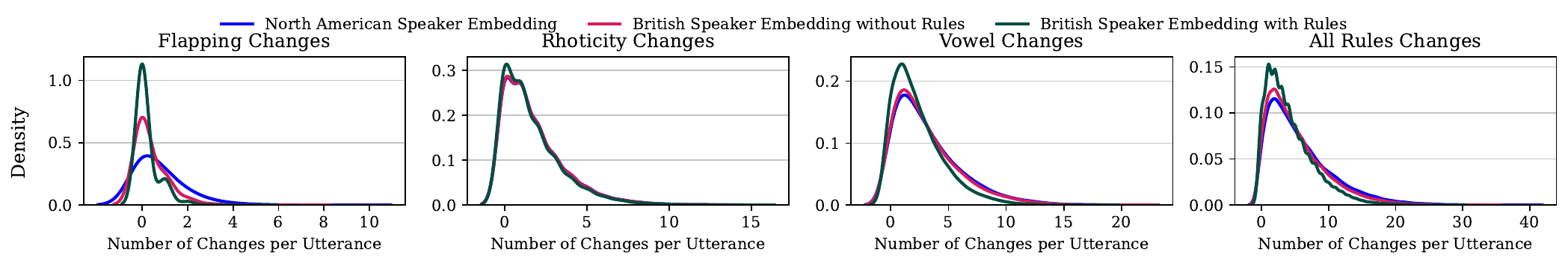}
    \vspace{-5mm}
    \caption{Kernel Density Estimation (KDE) of histograms on the number of changes of the rules.}
    \vspace{-5mm}
    \label{fig:histogram}
\end{figure*}

\subsection{Effects of Each Phonological Rule on Speech Synthesis}
\begin{table}[h!]
\vspace{-20pt}
\centering
\caption{Results on accented speech synthesis under different phonological transformation rules. All rules applied in this experiment are crafted for British English. We also present results conditioned on speaker embeddings from both accents to study how strongly these speaker embeddings adhere to or override these rule-driven transformations. ``NA'' denotes North American. ``B'' denotes British. ``Spk Emb'' denotes speaker embedding.}
\label{tab:analysis_of_each_rule}
\resizebox{\linewidth}{!}{
\begin{tabular}{l c cc cc c} 
\toprule
& \multirow{2}{*}{\textbf{UTMOS}} & \multicolumn{2}{c}{\textbf{Accent Prob}}  & \multicolumn{2}{c}{\textbf{Accent Sim}} & \multirow{2}{*}{\textbf{PSR}$\downarrow$}  \\
& & NA $\downarrow$ & B $\uparrow$ & NA $\downarrow$ & B $\uparrow$\\ 
\midrule 
\textbf{NA Spk Emb} & 4.43 & 86.5 & 3.79 & 0.85 & -0.05 & 0.856 \\ 
+ All & 4.42 & \textbf{58.8} & \textbf{17.3} & \textbf{0.74} & \textbf{0.21} & \textbf{0.827} \\
\midrule

\textbf{B Spk Emb} & 3.74 & 17.6 & 67.8 & 0.33 & 0.67 & 0.775 \\ 
+ Flapping & 3.74 & 17.3 & 68.5 & 0.32 & 0.67 & 0.749 \\
+ Rhoticity & 3.73 & 9.5 & 68.8 & 0.14 & 0.78 & 0.739 \\
+ Vowel & 3.73 & 9.8 & 77.8 & 0.18 & 0.78 & 0.693 \\
+ All & 3.72 & \textbf{5.3} & \textbf{78.4} & \textbf{0.03} & \textbf{0.85} & \textbf{0.628} \\
\hspace{0.25cm} - Flapping & 3.72 & 5.3 & 78.1 & 0.03 & 0.85 & 0.646 \\
\hspace{0.25cm} - Rhoticity & 3.73 & 9.5 & 78.1 & 0.18 & 0.78 & 0.670 \\
\hspace{0.25cm} - Vowel & 3.73 & 9.4 & 69.4 & 0.14 & 0.79 & 0.716 \\
\bottomrule 
\end{tabular}
}
\end{table}


Table~\ref{tab:analysis_of_each_rule} summarizes naturalness (UTMOS), accent strength (probabilities and similarities for North American vs. British), and phoneme shift rate (PSR) under different rule configurations. The column `Accent Prob' with `(NA/B)' indicates the average probabilities assigned by the Vox-Profile accent classifier to North American or British speech. The column `Accent Sim' with `(NA/B)' denotes the average cosine similarity between synthesized utterances and group-level reference embeddings for each accent. PSR measures how often rule-driven phoneme substitutions persist in the output, with lower values reflecting stronger rule preservation. 

Across conditions, \textbf{naturalness} is stable. UTMOS remains around 4.4 for North American and 3.7 for British settings, whether or not rules are applied. This indicates that integrating phonological rules alongside speaker embeddings does not degrade the naturalness of the synthesized speech. Consistently lower UTMOS scores across the board for the British-accented speech likely reflects bias in the predictor, which was trained on data with heavier North American representation~\cite{zhong2025pairwise} rather than any actual degradation.

Second, \textbf{accent probabilities} show clear rule effects. With a North American speaker embedding, the baseline system yields an 86.5\% probability of the North American accent. Adding all three phonological transformation rules drops this to 58.8\% and raises the British probability to 17.3\%, showing that rules shift the classifier's decision away from North American toward British. Conversely, with a British speaker embedding, the baseline system already achieves 67.8\% British probability, but this increases to 78.4\% when all British English substitution rules are applied.

Third, \textbf{accent similarities} follow the same trend. With North American embeddings, British similarity increases from -0.05 to 0.21 when rules are applied, pulling synthesized speech closer to the British reference. With British embeddings, similarity to the British reference rises from 0.67 to 0.85.

The \textbf{role of individual rules} emerges when applied selectively with British embeddings. Vowel correspondences drive the largest gains, raising British probability to 77.8\% and reducing PSR to 0.693. Rhoticity increases similarity to the British embedding space (0.78) even when classifier probabilities remain modest. Flapping alone has minimal impact but contributes additively when combined with other rules. Ablations confirm vowel correspondences as the single most influential factor.

Finally, PSR captures how often rule effects survive in output phonemes. Lower values indicate stronger preservation of rule-driven changes. With British embeddings, PSR falls from 0.775 without rules to 0.628 with all rules, showing that rules measurably shape outputs even when embeddings exert a strong pull in the opposite direction.

\subsection{How Do Rules Affect Phonemes of the Synthesized Speech?}

\begin{table}[h!]
\vspace{-20pt}
\centering
\caption{Total number of times we apply the rule (in thousands). $N1$ and $N2$ are the notation from Fig~\ref{fig:diagram} in the green highlight. NA and B denote North American and British accents.}
\label{tab:times_apply}
\footnotesize
\begin{tabular}{lcccc}
\toprule
 & \textbf{Flapping} & \textbf{Rhoticity} & \textbf{Vowel} & \textbf{All Rules} \\
\midrule
NA ($N1$) & 12.8 & 83.5 & 125.1 & 221.4 \\
NA ($N2$) & 25.3 & 57.9 & 106.3 & 189.5 \\
B w/o rules ($N2$) & 12.3 & 57.4 & 101.7 & 171.5 \\
B w/ rules ($N2$) & \textbf{6.7} & \textbf{53.7} & \textbf{78.5} & \textbf{139.0} \\
\bottomrule
\end{tabular}
\end{table}

To analyze rule retention at the phoneme level, we compare the number of substitutions ($N1$) expected with the number observed in recognized outputs ($N2$), as summarized in Table~\ref{tab:times_apply}. Ideally, if all substitutions survived, $N2$ would be zero after applying rules. It is often observed that speaker embeddings overwrite some substitutions, pushing outputs back toward their accent biases. 

The results show that vowel correspondences account for the largest number of changes, consistent with their strong effect in Table\ref{tab:analysis_of_each_rule}. For flapping, $N2$ increases under North American embeddings, suggesting that the G2P baseline phonemes were not fully American and the embedding reinforced flapping. For rhoticity and vowels, $N2$ is significantly lower than $N1$ under North American embeddings, demonstrating that one speaker embedding can simultaneously represent different accent tendencies depending on the phoneme context. 

Fig~\ref{fig:histogram} visualized these effects using KDE histograms; rule application increases the skewness toward smaller numbers of changes, indicating the effectiveness of the rules. However, distributions remain broad, highlighting how speaker embedding can override or dilute specific rule effects.

\vspace{-3mm}
\subsection{On the Role of Individual Speaker Embeddings in Accented Speech Synthesis.}

\begin{table}[h!]
\vspace{-20pt}
\centering
\caption{Results on accented speech synthesis under different
speaker embeddings. -R suffix denotes the addition of rules. NA and B denote North American and British accents.}
\label{tab:emb_comparison}
\footnotesize
\begin{tabular}{l cc cc c}
\toprule
& \multicolumn{2}{c}{\textbf{Accent Prob}}  & \multicolumn{2}{c}{\textbf{Accent Sim}} & \textbf{PSR} $\downarrow$ \\ 
& \textbf{NA $\downarrow$} & \textbf{B $\uparrow$} & \textbf{NA $\downarrow$} & \textbf{B $\uparrow$} & \\ 
\midrule 
Isabella & 2.7 & 85.3 & 0.03 & 0.88 & 0.633 \\
Isabella-R & \textbf{1.1} & \textbf{90.1} & \textbf{-0.07} & \textbf{0.92} & \textbf{0.481}\rlap{$_{\scriptsize -15.2\%}$} \\
\midrule
Lily & 2.0 & 89.3 & -0.01 & 0.91 & 0.660\\
Lily-R & \textbf{0.9} & \textbf{91.6} & \textbf{-0.12} &\textbf{ 0.93} & \textbf{0.494}\rlap{$_{\scriptsize -16.6\%}$} \\
\midrule
Fable & 17.6 & 67.8 & 0.33 & 0.67 & 0.775 \\ 
Fable-R & \textbf{5.7} & \textbf{78.4} & \textbf{0.03} & \textbf{0.86} & \textbf{0.628}\rlap{$_{\scriptsize -14.7\%}$} \\ 
\midrule
Daniel & 4.7 & 89.8 & 0.07 & 0.86 & 0.706 \\
Daniel-R & \textbf{1.5} & \textbf{93.2} & \textbf{-0.07} & \textbf{0.93} & \textbf{0.543}\rlap{$_{\scriptsize -16.3\%}$} \\
\bottomrule 
\end{tabular}
\end{table}

Table~\ref{tab:emb_comparison} examines how individual speaker embeddings interact with rule application. Results are shown for Kokoro TTS voices\footnote{https://huggingface.co/hexgrad/Kokoro-82M/blob/main/VOICES.md}. Across voices, rules consistently strengthen accent probabilities. For example, the Fable embedding alone yields 67.8\% British probability; with rules, this rises to 78.4\%. Similarity to the British reference also improves, and PSR decreases from 0.775 to 0.628. For Isabella and Lily, rule application produces 15-17\% absolute reductions in PSR, confirming stronger preservation of rule-driven transformations.

The degree of interaction varies across embeddings. Some embeddings like Daniel, already produce high British probabilities ($\sim89.8\%$) but still show improvements when rules are applied (to 93.2\%). Others, like Fable, rely more heavily on rule guidance. This variation suggests that embeddings encode accent characteristics with differing levels of entanglement, which rules can either reinforce or correct.

\vspace{-2mm}
\section{Conclusion and Future Work}
Taken together, our results show that phonological rules are a coarse but effective lever for accent control. They preserve naturalness, strengthen accent attributes, and reveal how speaker embeddings can reinforce or override explicit transformations. By offering interpretable, linguistically grounded modifications, rules complement speaker embeddings and highlight future directions for designing embeddings that disentangle accent from other traits while remaining responsive to rule-based inputs. However, as this study relies on automated metrics and is subject to the noise of the specific phoneme recognition model used, future work will incorporate a broader array of recognition architectures and human evaluations.  

\section{Acknowledgment}
This work was supported by the Office of the Director of National Intelligence (ODNI), Intelligence Advanced Research Projects Activity (IARPA), via the ARTS Program under contract D2023-2308110001. The views and conclusions contained herein are those of the authors and should not be interpreted as necessarily representing the official policies, either expressed or implied, of ODNI, IARPA, or the U.S. Government. The U.S. Government is authorized to reproduce and distribute reprints for governmental purposes notwithstanding any copyright annotation therein.

\bibliographystyle{IEEEbib}
\bibliography{strings,refs}

\end{document}